\documentclass{article}

\usepackage[utf8]{inputenc}
\usepackage{spconf}
\usepackage{mathtools}
\usepackage{tcolorbox}
\usepackage{amsmath}
\usepackage{amssymb}
\usepackage{amsfonts}
\usepackage{mathtools}
\usepackage{times}
\usepackage{tikz}
\usepackage{enumitem}
\usepackage{algorithm, algpseudocode}
\usepackage{multirow}

\input{mysymbol.sty}

\title{Matrix Low-Rank Approximation For Policy Gradient Methods}

\name{\vspace{-.25cm}Sergio Rozada, and Antonio G. Marques \thanks{ Work supported by the Spanish NSF Grants SPGraph (PID2019-105032GB-I00/AEI/10.13039/501100011033) and DATRASCOOP@SESM (TED2021-130347B-I00).
All the authors are with the Dept. of Signal Theory and Comms., King Juan Carlos University, Madrid, Spain. Email contact author: antonio.garcia.marques@urjc.es.} 
}
\address{Dept. of Signal Theory and Communications, King Juan Carlos University, Madrid, Spain}

\begin{document}
\maketitle

\ninept

\begin{abstract}%
Estimating a policy that maps states to actions is a central problem in reinforcement learning. Traditionally, policies are inferred from the so called value functions (VFs), but exact VF computation suffers from the curse of dimensionality. Policy gradient (PG) methods bypass this by learning directly a parametric \emph{stochastic} policy. Typically, the parameters of the policy are estimated using neural networks (NNs) tuned via stochastic gradient descent. However, finding adequate NN architectures can be challenging, and convergence issues are common as well. In this paper, we put forth \emph{low-rank} matrix-based models to estimate efficiently the parameters of PG algorithms. We collect the parameters of the stochastic policy into a matrix, and then, we leverage matrix-completion techniques to promote (enforce) \emph{low rank}. We demonstrate via numerical studies how \emph{low-rank} matrix-based policy models reduce the computational and sample complexities relative to NN models, while achieving a similar aggregated reward.
\end{abstract}



\section{Introduction} \label{S:Introduction}

In the era of big data, complex dynamical systems call for intelligent algorithms that use observations to adapt the way they behave with the world. Reinforcement Learning (RL) addresses this by implementing \emph{stochastic} algorithms that learn by trial and error how to interact with the environment \cite{sutton2018reinforcement, bertsekas2019reinforcement}. Technically speaking, RL models decision-making problems: given an environment defined by a set of states, the RL goal is to learn the best action to be made in each state. Then, RL is about estimating the reward associated with state-action pairs, which is referred to as value function (VF),  and  inferring a function that maps the states into actions, which is referred to as policy. Typically, RL algorithms estimate first the VF (strictly speaking, the expected accumulated reward associated with each state-action pair) and, then, implement a policy that maximizes the VF. However, value-based methods are algorithmically challenging, and suffer from the curse of dimensionality. To overcome these issues, policy-based methods were introduced. They bypass these problems by estimating the policy directly. To render the design of these methods tractable, policies are typically assumed stochastic and parametric, so that the problem boils down to estimating the parameters of some probability distribution.

The workhorse approach in policy approaches is to consider normal distributions and learn the parameters of those Gaussians using a neural network (NN) architecture that uses the states as inputs \cite{arulkumaran2017deep}. NNs, and therefore, most policy-based methods tend to suffer from convergence issues, as they strongly depend on the architecture employed. As a result, postulating (finding) the proper NN architecture for the RL setup at hand is usually a problem in itself.

In the context of stochastic policy methods for RL, this paper takes an alternative path and, leveraging matrix completion results \cite{eckart1936approximation,markovsky2012low}, puts forth a low-rank algorithm to estimate stochastic policy models. The ultimate goal is to design an estimation scheme that i) is sufficiently generic to learn the policy and ii) mitigates some of the problems present in NN-based schemes. 
The rest of this section is devoted to explaining the state of the art and the contribution in more detail. Fundamentals of RL are discussed in Section 2, and our new low-rank scheme is 
introduced in Section 3. Numerical experiments showcasing some of the benefits of our approach are provided in Section 4.

\vspace{.05cm}
\noindent \textbf{Detailed contribution and related work.} This paper investigates the design of \emph{low-rank} schemes for (actor/critic) policy RL methods leveraging matrix completion techniques. The central aspects of such a design are i) the consideration of a policy method based on an actor and a critic (where the latter estimates a VF); ii) modeling the actor parameters and the critic VF as matrices (typically associated with a clusterization of the different dimensions of the input state); and iii) to regularize the estimation problem by enforcing a low-rank structure via matrix factorization. Although those techniques are well-known in the context of low-rank optimization and matrix completion \cite{eckart1936approximation,markovsky2012low,udell2016generalized,mardani2013decentralized}, they have not been exploited in the context of policy-based RL methods. While closely-related ways of modeling parsimony in RL have been investigated, e.g. sparsity \cite{tolstaya2018nonparametric, lever2016compressed}, the use of low-rank optimization in the general context of RL has been limited. Notable exceptions include efforts to approximate some structures of the Markov Decision Process (MDP), such as the transition matrix or the reward function, as low rank \cite{barreto2016incremental, jiang2017contextual, mahajan2021tesseract}. Matrix-completion techniques have also been used to compress the estimated $Q$-functions of an MDP \cite{ong2015value}. In energy storage applications, low-rank (rank-one) methods have been proposed for the estimation of the VF \cite{cheng2016co,cheng2017low}. Along the same lines, \emph{linear models} have been used to approximate VFs on-the-fly \cite{melo2007q}. These methods operate by defining a set of features \cite{behzadian2019fast} and then modeling the VF as a weighted sum of the features associated with the state-action pair. More recently, in the context of value-based methods, schemes that estimate the VF promoting a low-rank matrix structure \cite{yang2019harnessing, shah2020sample, rozada2021low} and a low-rank tensor structure \cite{rozada2022tensor} have been proposed.


\section{Fundamentals of RL and notation.}

RL models the world as a closed-loop setup where agent(s) sequentially interact with the environment. The environment consists of i) the space of states $\ccalS$; ii) the space of action agent(s) can take $\ccalA$; and iii) the reward associated with every state-action pair. In each (typically multidimensional) state, the agent takes an action and obtains a reward. 
To be more specific, let $t=1,...,T$ be a time index. Given a particular state $s_t$, the agent takes an action $a_t$, and obtains a reward $r_t$. The reward $r_t$ quantifies the \emph{instantaneous} value of the state-action pair. However, the action $a_t$ impacts $s_{t'}$ for $t'>t$ and, as a result, affects future rewards $r_{t'}$ for $t'>t$. This illustrates that actions are coupled across time and that the aggregated (long-term) reward must be considered for decision-making. Furthermore, the dependence of $r_t$ on $s_t$ and $a_t$ is usually stochastic, rendering the optimization more challenging. Markovianity is commonly used to mitigate these issues, so that the optimization can be recast as a MDP.

In this context, i) a policy $\pi:\ccalS\mapsto\ccalA$ is a function that maps states into actions; and ii) the VF associated with $(s_t,a_t)$ is the expected  cumulative reward $\mathbb{E}[\sum_{t'=t}^{T}\gamma^{t'-t}r_{t'}|s_t,a_t]$, with $\gamma\in(0,1)$ being a discount factor that places more focus on short-term rewards \cite{sutton2018reinforcement,bertsekas2019reinforcement}. For a specific policy $\pi$, the VF quantifies the expected reward associated with a particular state by taking the actions dictated by $\pi$. Value-based methods focus on estimating the VFs (using parametric or non-parametric approaches) of the MDP, to infer a policy later on from the VFs by greedily following the path of highest value \cite{watkins1992q, mnih2013playing}. Policy-based methods, on the other hand, focus on directly learning the policy $\pi$, typically using a stochastic parametric models \cite{williams1992simple, peters2008natural}. To estimate the parameters, a set of trajectories sampled from the environment is assumed to be available, and those are used to run a stochastic gradient descent scheme. Each trajectory $\tau_T:(s_0, a_0, r_0),...,(s_T, a_T, r_T)$ is a set of state-action-rewards triplets obtained from the sequential interaction of an agent with the environment. 

From an optimization perspective, policy-based methods aim at maximizing the reward of a trajectory $\ccalR(\tau_T)$ over all possible trajectories $\tau$. To be precise, consider the following objective function:

\begin{equation}
    \label{eq::pg_cost}
    J(\theta) = \mathbb{E}_{\ccalT_\theta(\tau_T)} [\ccalR(\tau_T)]
\end{equation}

\noindent where $\theta$ denotes the parameters of a policy $\pi_\theta$. Then, $\ccalT_\theta(\cdot)$ is the probability density function of the trajectories under the given policy $\pi_\theta$. Leveraging Markovianity, it follows that $\ccalT_\theta (\tau_T) = \ccalD(s_0) \prod_{t=0}^T \pi_\theta(a_t | s_t) P(s_{t+1} | s_t, a_t)$, where $\ccalD(\cdot)$ is the initial distribution of the MDP. Since direct optimization of     \eqref{eq::pg_cost} is intractable, the standard approach is to estimate the parameters $\theta$ of the policy $\pi_\theta$ via (stochastic) gradient ascent methods \cite{lee2020optimization}. The update rule takes the form $\theta_{h+1} = \theta_h + \alpha_\theta \nabla_\theta J(\theta_h)$, where $h$ is the iteration index, and $\alpha_\theta$ is the gradient step, or learning rate. Obtaining the gradient of $J$ is challenging since it involves computing derivatives of an expectation across $\ccalT_\theta$. The policy gradient theorem \cite{sutton1999policy} provides an elegant reformulation of $\nabla_\theta J(\theta)$
\begin{equation}
    \label{eq::pg_gradient}
    \nabla_\theta J(\theta) = \mathbb{E}_{\ccalT_\theta} [\sum_{t=0}^T \ccalR(\tau_T) \nabla_\theta log\;\pi_\theta(a_t | s_t)],
\end{equation}
which is more tractable. 
One of the most celebrated policy-based algorithms is REINFORCE \cite{williams1992simple}, or Monte-Carlo policy gradient, which samples a full trajectory from the environment to avoid computing the expectation, and uses \eqref{eq::pg_gradient} to update the parameters using the actual total return $G_t=\sum_{t'=t}^T r_{t'}$ in lieu of $\ccalR(\tau_T)$. 

\vspace{.5mm}
\noindent\textbf{Actor critic methods.} The main drawback of Monte-Carlo methods is that they exhibit a high variance, due to so called credit assignment problem. This issue is usually addressed by subtracting a baseline value $b_t$ from the return $G_t$ to reduce the variance of the gradient estimate while keeping the bias unchanged \cite{greensmith2004variance}. A choice that provides several benefits is setting $b_t$ to the VF. The quantity resultant from this  subtraction is denoted as $A(s_t) = G_t - V(s_t)$ and is called advantage function. The price to pay in this case is that a method to estimate the VFs is needed as well. This leads to a two-step method (actor-critic) that involves the estimation of the parameters of the policy $\pi_\theta$ (actor) as well as those of the VF (critic). The parameters of the actor $\theta$ are updated via gradient ascent using
\begin{equation}
    \label{eq::pg_ac_gradient}
    \nabla_\theta J(\theta) = \sum_{t=0}^T A(s_t) \nabla_\theta log\;\pi_\theta(a_t | s_t),
\end{equation}
where $A(s_t) = G_t - V_\omega(s_t)$, and $V_\omega(s_t)$ is considered given. In contrast, the parameters $\omega$ of the VF are obtained as 
\begin{equation}
    \label{eq::critic_loss}
    \omega^*=\arg\min_\omega \ccalL(\omega) := \arg\min_\omega \frac{1}{2}\sum_{t=0}^T (G_t - V_\omega(s_t))^2.
\end{equation}
While some parametric models \eqref{eq::critic_loss} can be solved in closed form,  $\omega$ are typically estimated using gradient descent methods of the form 
\begin{equation}
\omega_{h+1}=\omega_{h} - \alpha_\omega \nabla_\omega \ccalL(\omega_h),
\end{equation} 
where $h$ is the iteration index and $\alpha_\omega$ is the gradient step.

\vspace{1mm}
\noindent\textbf{Gaussian policies.}
In setups where the action state $\ccalA$ is continuous and one-dimensional, a commonly adopted approach is to model actions as samples of a univariate Gaussian distribution $a_t\sim \ccalN(a | \mu(s_t), \sigma(s_t))$  \cite{levine2014learning, ciosek2018expected}. The problem of finding the optimal action associated with a state is reformulated as finding the mean $\mu(s_t)$ and the standard deviation $\sigma(s_t)$ associated with the state. The functions $\mu:\ccalS \mapsto \reals$ and $\sigma:\ccalS \mapsto \reals^+$ can be either parametric or non-parametric, with a stronger preference in the literature for the former. Clearly, to find the gradient on the right-hand side of    \eqref{eq::pg_ac_gradient}
 w.r.t. those parameters, one needs to resort to the chain rule and find first the derivative of $J$ w.r.t. $\mu$ and $\sigma$. Upon setting the distribution of the policy to a Gaussian, this approach leads to the following partial derivatives
\begin{align}
	\nabla_\mu J(\mu, \sigma) &= \sum_{t=0}^T A(s_t) \frac{a_t - \mu(s_t)}{\sigma(s_t)^2}, \label{eq::pg_ac_gradient_mu} \\
	\nabla_\sigma J(\mu, \sigma) &= \sum_{t=0}^T A(s_t) \left( \frac{(a_t - \mu(s_t))^2}{2\sigma(s_t)^3} - \frac{1}{\sigma(s_t)}\right). \label{eq::pg_ac_gradient_sigma}
\end{align}
\noindent Practical implementations often consider simpler models for $\sigma(s_t)$. For example, it is not uncommon to assume that $\sigma$ is constant and independent of the state $s_t$, reducing the problem to learn this single value (or even fixing it using prior knowledge).

After describing the basics of policy methods in RL, we are ready to introduce our scheme. The reader should notice that al- though in this paper we have focused on Gaussian policies, the proposed approach can be extended to other policy models.


\section{Policy Gradients and Matrix Factorization}
\label{S:pgmf}

The state space $\ccalS$ in most MDPs is discrete, either because the number of states is finite or because it represents the discretized (sampled) version of a continuous space. As a result, we have a finite number $N_\ccalS$ of states. Since each state is mapped to a mean and standard deviation, the goal in this section is to provide a (low-rank based) scheme that yields an efficient estimator for the $N_\ccalS$  mean and standard deviation pairs. 

The first step in our approach is to introduce a matrix representation of the states. In most applications states are multi-dimensional and as a result, they can be indexed by a tuple of indexes. To keep the exposition simple, suppose that two indices are used (either because there are only two dimensions, or because the different dimensions are clustered into two groups)\footnote{Generalizations of this approach to (low-rank) tensor models are straighforward and well-motivated, but for simplicity, we limit this conference paper to the matrix case.}. With this in mind, define $\bbX_\mu\in\reals^{N\times M}$ as the matrix that collects the means associated with each of the $N_\ccalS=N\cdot M$ states. The idea under this approach is that every state $s \in \ccalS$ is coded into two indices $i_s \in \{1,...,N\}$ and $j_s \in \{1,...,M\}$ and, then, the mean associated with $s_t$ is simply obtained as $\mu (s_t) = [\bbX_\mu]_{i_{s_t}, j_{s_t}}$. Analogously, we define $\bbX_\sigma\in\reals^{N\times M}$ and set $\sigma (s_t) = [\bbX_\sigma]_{i_{s_t}, j_{s_t}}$. If no structure is imposed on $\bbX_\mu$ and $\bbX_\sigma$, the estimation of $\mu$ and $\sigma$ is non-parametric and, as a result, the  updates in \eqref{eq::pg_ac_gradient_mu} and \eqref{eq::pg_ac_gradient_sigma} suffice to estimate the policy.

However, such an approach would suffer from the curse of dimensionality. To avoid this, we impose additional structure on $\bbX_\mu$ and $\bbX_\sigma$, forcing them to be low rank\footnote{Alternative schemes that \emph{promote} low-rank via, e.g., nuclear norm regularizers are also possible.}. In particular, this entails introducing the matrices $\bbL_\mu \in \reals^{M \times K} $, $\bbR_\mu  \in \reals^{K \times N}$, $\bbL_\sigma \in \reals^{M \times K} $ and $\bbR_\sigma  \in \reals^{K \times N}$, and write the mean and standard deviation matrices as $\bbX_\mu=\bbL_\mu \bbR_\mu$ and $\bbX_\sigma=\bbL_\sigma \bbR_\sigma$. This model is still non-parametric, but it helps alleviating the curse of dimensionality. The implications of this design decision are twofold: i) the rank of $\bbX_\mu$ and $\bbX_\sigma$ is at most $K$, limiting the degrees of freedom of those matrices (hence, facilitating its estimation from a limited number of observations); and ii) the parameters to estimate are no longer the entries of $\bbX_\mu$ and  $\bbX_\sigma$, but the entries of $\bbL_\mu$, $\bbR_\mu$, $\bbL_\sigma$ and $\bbR_\sigma$. To see the latter point more clearly, note that the mapping from the state to the parameters of the Gaussian under the low-rank models is given by 
\begin{equation}\label{eq:mu_sigma_polynomial_order_two}
\mu (s_t\!) \!= \!\!\sum_{k=1}^K [\bbL_\mu]_{i_{s_t}\!, k} [\bbR_\mu]_{k, j_{s_t}}\;|\;\sigma (s_t\!) \!= \!\!\sum_{k=1}^K [\bbL_\sigma]_{i_{s_t}\!, k} [\bbR_\sigma]_{k, j_{s_t}}\!.
\end{equation}

The next step is to optimize/estimate the values of the entries of  $\bbL_\mu$, $\bbR_\mu$, $\bbL_\sigma$ and, $\bbR_\sigma$. To that end, we resort to non-convex gradient-based matrix factorization approaches \cite{markovsky2012low} along with the expressions for the gradients provided in \eqref{eq::pg_ac_gradient_mu}-\eqref{eq::pg_ac_gradient_sigma}.  
To be more specific, in our approach the parameters $\theta$ are the entries of 	$\{\bbL_\mu, \bbR_\mu, \bbL_\sigma, \bbR_\sigma\}$ and our goal is to find an expression for the (entries of the) gradient in     \eqref{eq::pg_ac_gradient}. For simplicity, let us focus first on $\bbL_\mu$ and $\bbR_\mu$. We aim at finding the expression for the entries of $\nabla_{\bbL_\mu} J(\bbL_\mu, \bbR_\mu, \bbL_\sigma, \bbR_\sigma)$, and  $\nabla_{\bbR_\mu} J(\bbL_\mu, \bbR_\mu, \bbL_\sigma, \bbR_\sigma)$. To that end, we need to apply the chain rule and combine the partial derivatives of the cost $J$ w.r.t. $\mu$ in \eqref{eq::pg_ac_gradient_mu}
with the partial derivatives of the $\mu$ function in   \eqref{eq:mu_sigma_polynomial_order_two} w.r.t. each entry of the matrices $\bbL_\mu$, and $\bbR_\mu$. The result of this is
\begin{align}
    \frac{\partial J(\bbL_\mu, \bbR_\mu, \bbL_\sigma, \bbR_\sigma)}{\partial [\bbL_\mu]_{i, k}} = 
    \sum_{t=0}^T \mathbb{I}_{i=i_{s_t}} A(s_t) \nonumber \\ \frac{a_t - [\bbL_\mu \bbR_\mu]_{i_{s_t}, j_{s_t}}}{[\bbL_\sigma \bbR_\sigma]_{i_{s_t}, j_{s_t}}^2} [\bbR_\mu]_{k, j_{s_t}} \label{eq::pg_l_mu} \\
    \frac{\partial J(\bbL_\mu, \bbR_\mu, \bbL_\sigma, \bbR_\sigma)}{\partial [\bbR_\mu]_{k, j}} = 
    \sum_{t=0}^T \mathbb{I}_{j=j_{s_t}} A(s_t) \nonumber \\
    \frac{a_t - [\bbL_\mu \bbR_\mu]_{i_{s_t}, j_{s_t}}}{[\bbL_\sigma \bbR_\sigma]_{i_{s_t}, j_{s_t}}^2} [\bbL_\mu]_{i_{s_t}, k} \label{eq::pg_r_mu} 
\end{align}
\noindent where $\mathbb{I}_{i=i_{s_t}}$, and $\mathbb{I}_{j=j_{s_t}}$ are indicator functions. An analogous approach considering the derivatives of the cost $J$ w.r.t. $\sigma$ in \eqref{eq::pg_ac_gradient_mu}, 
along with the derivatives of  $\sigma$ w.r.t. the entries of $\bbL_\sigma$ and $\bbR_\sigma$ yields:
\begin{align}
    \frac{\partial J(\bbL_\mu, \bbR_\mu, \bbL_\sigma, \bbR_\sigma)}{\partial [\bbL_\sigma]_{i, k}} =
    \sum_{t=0}^T \mathbb{I}_{i=i_{s_t}} A(s_t) \nonumber \\
    \left( \frac{(a_t - [\bbL_\mu \bbR_\mu]_{i_{s_t}, j_{s_t}})^2}{2[\bbL_\sigma \bbR_\sigma]_{i_{s_t}, j_{s_t}}^3} - \frac{1}{[\bbL_\sigma \bbR_\sigma]_{i_{s_t}, j_{s_t}}}\right) [\bbR_\sigma]_{k, j_{s_t}} \label{eq::pg_l_sigma}
\end{align}
\begin{align}
   \frac{\partial J(\bbL_\mu, \bbR_\mu, \bbL_\sigma, \bbR_\sigma)}{\partial [\bbR_\sigma]_{k, j}} =
    \sum_{t=0}^T \mathbb{I}_{j=j_{s_t}} A(s_t) \nonumber \\
    \left( \frac{(a_t - [\bbL_\mu \bbR_\mu]_{i_{s_t}, j_{s_t}})^2}{2[\bbL_\sigma \bbR_\sigma]_{i_{s_t}, j_{s_t}}^3} - \frac{1}{[\bbL_\sigma \bbR_\sigma]_{i_{s_t}, j_{s_t}}}\right) [\bbL_\sigma]_{i_{s_t}, k} \label{eq::pg_r_sigma} 
\end{align}

Once the actor is properly set up, the last step is to define the scheme to estimate $\omega$, the parameters of the critic. As for the actor, we use a matrix representation $\bbX_\omega \!\in\! \reals^{N\times M}$ for the VF and then postulate that the VF matrix is low rank, so that it can be factorized as the product of $\bbL_\omega \!\in \! \reals^{N\times K}$ and $\bbR_\omega \!\in\! \reals^{K\times M}$ with $K\ll$ $\min\{N,M\}$. As a result,  the functional expression of the critic takes the form $V(s_t) \!= \!\!\sum_{k=1}^K [\bbL_\omega]_{i_{s_t}\!, k} [\bbR_\omega]_{k, j_{s_t}\!}$. We then optimize the critic using alternating gradient descent, with the partial derivatives of the critic cost in  \eqref{eq::critic_loss} w.r.t. the entries of ${\bbL_\omega}$ and  $\bbR_\omega$ being
\begin{align}
    \frac{\partial \ccalL(\bbL_\omega, \bbR_\omega)}{\partial [\bbL_\omega]_{i, k}} = \sum_{t=0}^T \mathbb{I}_{i=i_{s_t}} (G_t - [\bbL_\omega \bbR_\omega]_{i_{s_t}, j_{s_t}})[\bbR_\omega]_{k, j_{s_t}} \label{eq::pg_l_omega} \\
    \frac{\partial \ccalL(\bbL_\omega, \bbR_\omega)}{\partial [\bbR_\omega]_{k, j}} = \sum_{t=0}^T \mathbb{I}_{j=j_{s_t}} (G_t - [\bbL_\omega \bbR_\omega]_{i_{s_t}, j_{s_t}})[\bbL_\omega]_{i_{s_t}, k} \label{eq::pg_r_omega}
\end{align}

\vspace{1mm}
\noindent\textbf{The algorithm.} A low-rank policy gradient (LRPG) algorithm flows from the definitions above. The agent samples a full trajectory using the Gaussian policy $\pi_{\mu, \sigma}=\ccalN(a_t | \mu(s_t), \sigma(s_t))$, where the mean is $\mu(s_t)=[\bbL_\mu \bbR_\mu]_{i_{s_t}, j_{s_t}}$, and the standard deviation is $\sigma(s_t)=[\bbL_\sigma \bbR_\sigma]_{i_{s_t}, j_{s_t}}$. Then, the actor matrices $\bbL_\mu$, $\bbR_\mu$, $\bbL_\sigma$, and $\bbR_\sigma$ are updated via stochastic gradient ascent using \eqref{eq::pg_l_mu}--\eqref{eq::pg_r_sigma}. Finally, the critic matrices $\bbL_\omega$, and $\bbR_\omega$ are updated via stochastic gradient descent using \eqref{eq::pg_l_omega}--\eqref{eq::pg_r_omega}. The algorithm is depicted in Algorithm  \ref{alg:pg}. 

\begin{algorithm}[!htbp]
\flushleft
\caption{Low Rank Policy Gradient (LRPG)}\label{alg:pg}
\begin{algorithmic}
    \Require Initial policy and VF matrices $\bbL_\mu^0, \bbR_\mu^0, \bbL_\sigma^0$, $\bbR_\sigma^0$,  $\bbL_\omega^0$, and $\bbR_\omega^0$; learning rates $\alpha_\mu$, $\alpha_\sigma$, 
    $\alpha_\omega$; and maximum number of episodes $H$.
    \For{$h=0, ...,  H$}
        \State{Observe initial state $s_0$}
        \For{$t=0, ..., T$} \hfill\Comment{Sample a trajectory $\tau_T$}
            \State{$\mu_{s_t} \gets [\bbL_\mu \bbR_\mu]_{i_{s_t}, j_{s_t}}$}
            \State{$\sigma_{s_t} \gets [\bbL_\sigma \bbR_\sigma]_{i_{s_t}, j_{s_t}}$}
            \State{$a_t \sim \ccalN(a|\mu_{s_t}, \sigma_{s_t})$}
            \State{Take action $a_t$, and observe next state $s_{t'}$, and reward $r_t$}
            \State{$s_t \gets s_{t'}$}
        \EndFor
        \State
        \State $\bbL_\mu^{h+1} \gets \bbL_\mu^h + \alpha_\mu \nabla_{\bbL_\mu} J(\bbL_\mu^h, \bbR_\mu^h, \bbL_\sigma^h, \bbR_\sigma^h)$ \hfill\Comment{Actor update}
        
        \State $\bbR_\mu^{h+1} \gets \bbR_\mu^h + \alpha_\mu \nabla_{\bbR_\mu} J(\bbL_\mu^h, \bbR_\mu^h, \bbL_\sigma^h, \bbR_\sigma^h)$ 
        
        \State $\bbL_\sigma^{h+1} \gets \bbL_\sigma^h + \alpha_\sigma \nabla_{\bbL_\sigma} J(\bbL_\mu^h, \bbR_\mu^h, \bbL_\sigma^h, \bbR_\sigma^h)$ 
        
        \State $\bbR_\sigma^{h+1} \gets \bbR_\sigma^h + \alpha_\sigma \nabla_{\bbR_\sigma} J(\bbL_\mu^h, \bbR_\mu^h, \bbL_\sigma^h, \bbR_\sigma^h)$ 
        
        \State
        \State $\bbL_\omega^{h+1} \gets \bbL_\omega^h + \alpha_\omega \nabla_{\bbL_\omega} J(\bbL_\omega^h, \bbR_\omega^h)$ \hfill\Comment{Critic update}

        \State $\bbR_\omega^{h+1} \gets \bbR_\omega^h + \alpha_\omega \nabla_{\bbR_\omega} J(\bbL_\omega^h, \bbR_\omega^h)$ 
    \EndFor
\end{algorithmic}
\end{algorithm}



\section{Numerical experiments}
\label{S:simulations}

We test LRPG in three standard continuous-action RL problems of the toolkit OpenAI Gym \cite{brockman2016openai}. The first scenario is the inverted pendulum, where an agent tries to keep a pendulum upright. The second one is the acrobot, which mimics a gymnast trying to swing up. The third one is the Goddard problem, a rocket optimizing its peak altitude ascending vertically. We test and compare NN-based vs low-rank-based policy methods. The figures of merit reported are i) parametrization-efficiency, ii) convergence rate, and iii) return obtained by the estimated policy. Details can be found in \cite{rozada2022code}.

\begin{figure}[h]
    \centering
    \includegraphics[width=.9\linewidth]{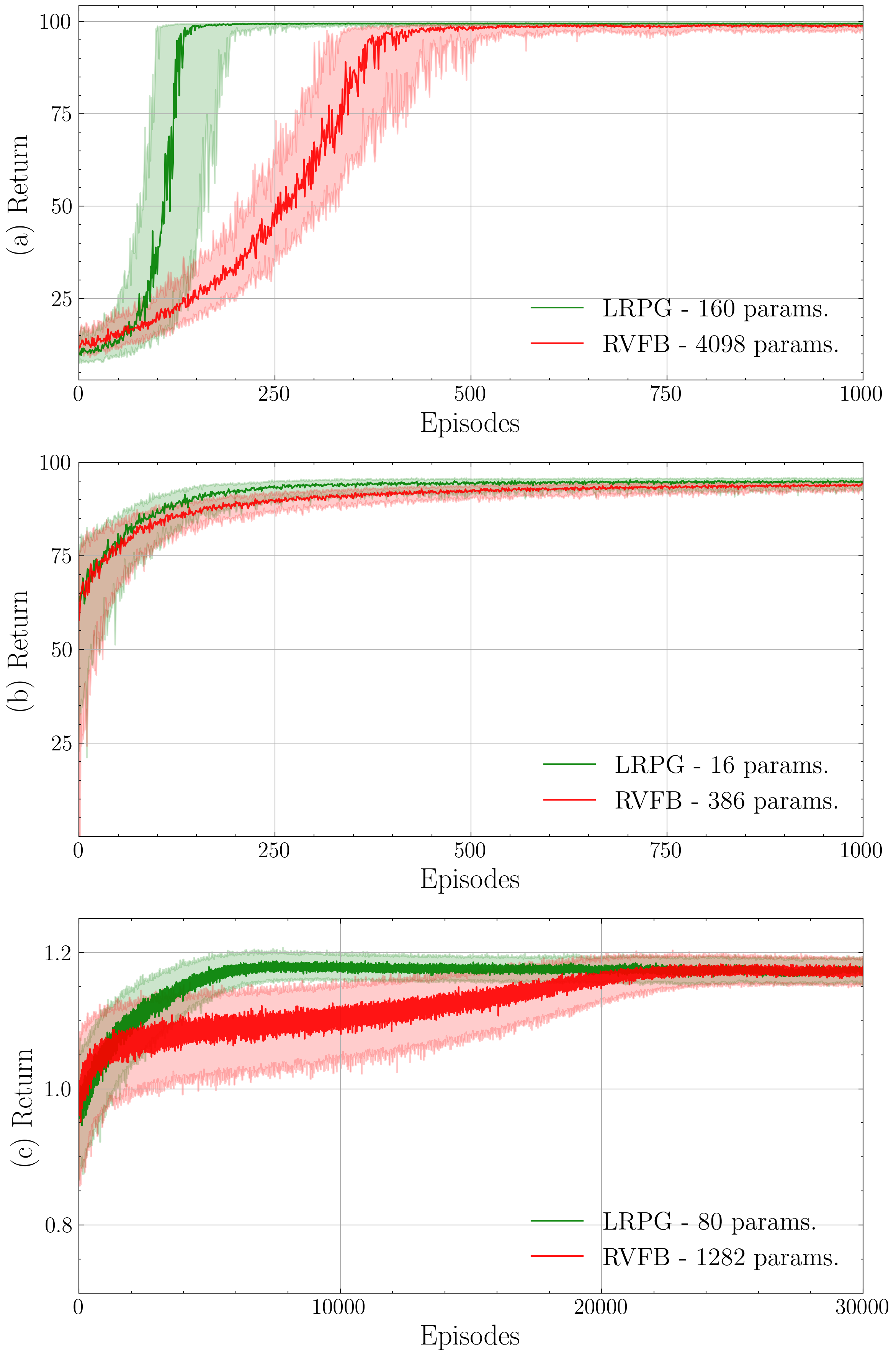}
    
    \vspace{-3mm}
    \caption{Median return per episode in three standard RL problems: (a) the pendulum, (b) the acrobot, and (c) the Goddard problem.}
    \label{fig::cumreward}
    \vspace{-5mm}
\end{figure}

\noindent \textbf{Experimental setup.} We compared the LRPG algorithm against REINFORCE with VF-baselines (RVFB). In both scenarios, given a state $s_t$, the action $a_t$ is sampled from the normal distribution $\ccalN(a|\mu(s_t), \sigma)$. For the sake of simplicity, $\sigma$ does not depend on $s_t$ and the difference between the schemes resides in the model to estimate the mean $\mu(s_t)$. While the parameters for LRPG are the entries of $\bbL_\mu$ and $\bbR_\mu$, the RVFB algorithm uses an NN model $\mu_\theta(s_t)=\text{NN}_\theta(s_t)$. Similarly, LRPG models the VF as $V(s_t)=[\bbL_\omega \bbR_\omega]_{i_{s_t}, j_{s_t}}$, and RVFB as $V_\omega(s_t)=\text{NN}_\omega(s_t)$. Since the three setups tested are continuous, LRPG discretizes the state space $\ccalS$. Note that the finer the discretization, the larger the number of entries of $\bbX_\mu$ and $\bbX_\omega$. Imposing low rank can drastically reduce the number of parameters while keeping a fine sampling resolution. The number of total states is defined by the Cartesian product of a regularly-sampled grid. On the other hand, NNs can deal with continuous setups, but choosing the proper architecture is usually a hard task. For the presented scenarios, we run $100$ simulations of each algorithm. In each episode, we measure the return per episode $\check{\ccalR}=\sum_{t=0}^T r_t$.  In Fig. \ref{fig::cumreward} the median $\check{\ccalR}$ (across the $100$ simulations) is shown. We have exhaustively examined the space of all potential fully-connected NN architectures to find the smallest one that solves the problem. A summary of the parametrizations and associated median returns is provided in Table \ref{tab::parameters-table}.

\begin{table}[]
{\footnotesize	
\begin{tabular}{l||ll|ll}
\multirow{2}{*}{\textbf{Env.}} & \multicolumn{2}{l}{\textbf{LRPG}} & \multicolumn{2}{l}{\textbf{RVFB}} \\
                             & \textbf{Parameters}    & \textbf{Return}   & \textbf{Parameters}    & \textbf{Return}   \\ \hline
Pendulum                     & 160            & \textbf{99.34}       & 4,098            & 98.81      \\ 
Acrobot                      & 16             & \textbf{94.45}        & 386              & 94.25       \\
Rocket                       & 80             & \textbf{1.17}        & 1,282            & 1.16      
\end{tabular}
\hfill \break
}
\vspace{-3mm}
\caption{Parameters vs. Median return.}
\label{tab::parameters-table}
\vspace{-4mm}
\end{table}

\noindent \textbf{Convergence properties.} The rate of convergence of LRPG and RVFB can be observed in Fig. \ref{fig::cumreward}. Here, we use the term convergence loosely to refer to the fact that the return reaches a steady state. In all scenarios, LRPG needs fewer episodes to converge to the final policy than RVFB. The strength of this finding  varies across scenarios, being conspicuous  in some of them. In the Pendulum scenario, LRPG algorithm needs almost $500$ episodes less to converge than RVFB (5 times faster). Similarly, in the Goddard problem, LRPG converges around episode $7,000$, while RVFB converges in episode $22,000$. Note, however, that this is not the case in the acrobot setup, where LRPG converges only slightly faster than RVFB. The faster convergence of LRPG is likely due to the fact that low-rank factorized models are simpler and have fewer parameters to estimate (see Table \ref{tab::parameters-table}). Furthermore, trading off between convergence rate and obtained return is significantly harder in NN-based setups since exploring and testing NN architectures is challenging.

\vspace{0.5mm}
\noindent \textbf{Parametrization efficiency.} As summarized in Table \ref{tab::parameters-table}, LRPG needs far fewer parameters than RVFB to estimate its associated policy. In the pendulum and acrobot environments, the size of the LRPG low-rank matrix-based models is approximately $~4\%$ of that of the NNs in RVFB. In the case of the Goddard problem, the size of the matrix model is around $~6\%$ of the size of the NN model. As mentioned previously, no smaller fully-connected NN was found to converge in these problems. Yet the policies estimated by LRPG lead to higher returns in all scenarios. NNs tend to have problems with local optima, especially in RL setups, and the space of all possible architectures is vast. Hence, the problem of finding the adequate architecture not only affects the reward, but also the convergence properties of the algorithm. In contrast, tuning the LRPG algorithm boils down to designing the discretization and indexing of the state space $\ccalS$, along with  the rank of the model $K$.



\section{Conclusions}
\label{S:conclusions}

This paper introduced a \emph{low-rank} policy gradient (LRPG) algorithm, which leverages matrix completion in (discrete) matrix-based Gaussian policy models. This approach portrays a natural way of introducing parsimony in policy-based RL techniques. The low-rank policy is very efficient in terms of parameters, and it is easy to interpret. LRPG  was tested in three OpenAI Gym classical control environments, comparing it with standard NN-based schemes. The limited but meaningful tests reveal that LRPG achieves high rewards, usually better than NN-based models, with a relatively small number of parameters. Furthermore, it requires significantly less number of episodes to converge to the final policy. 


\newpage
\bibliographystyle{IEEEtran}
\bibliography{references}

\end{document}